\documentclass[RESEARCH]{fcs}
\usepackage{hyphenat}
\usepackage{balance} 
\usepackage{multirow}
\usepackage{algorithm}
\usepackage{color}
\usepackage{colortbl}
\usepackage{algorithmic}
\usepackage{bm}

\usepackage{subfigure}
\usepackage{array}
\newcommand{\PreserveBackslash}[1]{\let\temp=\\#1\let\\=\temp}

\newcommand{\sol}{{\bm{\theta}}}
\newcommand{\data}{{\bm{\tau}}}
\newcommand{\x}{{\bm{x}}}   
\newcommand{\bd}{{\bm{d}}}
\newcommand{\bdspace}{{\mathcal{S}}}
\newcommand{\param}{{\bm{\phi}}}
\newcommand{\Prob}{P}
\newcommand{\simfn}{{\text{sim}}}
\usepackage{booktabs}

\title{Diversity from Human Feedback}

\author[1, 2]{Ren-Jian WANG$^*$}
\author[1, 2]{Ke XUE$^*$}
\author[1, 2]{Yutong WANG}
\author[3]{Peng YANG}
\author[4]{Haobo FU}
\author[4]{Qiang FU}
\author*[1, 2]{Chao QIAN}
\address[1]{National Key Laboratory for Novel Software Technology, Nanjing University, Nanjing 210023, China}
\address[2]{School of Artificial Intelligence, Nanjing University, Nanjing 210023, China}
\address[3]{Department of Statistics and Data Science, Southern University of Science and Technology, Shenzhen 518055, China}
\address[4]{Tencent AI Lab, Shenzhen 518054, China \\ $^*$ These authors contributed equally to this work
}
\fcssetup{
  received       = {month dd, yyyy},
  accepted       = {month dd, yyyy},
  corr-email     = {qianc@nju.edu.cn},  
}

\begin{abstract}
Diversity plays a significant role in many problems, such as ensemble learning, reinforcement learning, and combinatorial optimization. How to define the diversity measure is a longstanding problem. Many methods rely on expert experience to define a proper behavior space and then obtain the diversity measure, which is, however, challenging in many scenarios. In this paper, we propose the problem of learning a behavior space from human feedback and present a general method called Diversity from Human Feedback (DivHF) to solve it. DivHF learns a behavior descriptor consistent with human preference by querying human feedback. The learned behavior descriptor can be combined with any distance measure to define a diversity measure. We demonstrate the effectiveness of DivHF by integrating it with the Quality-Diversity optimization algorithm MAP-Elites and conducting experiments on the QDax suite. The results show that the behavior learned by DivHF is much more consistent with human requirements than the one learned by direct data-driven approaches without human feedback, and makes the final solutions more diverse under human preference. Our contributions include formulating the problem, proposing the DivHF method, and demonstrating its effectiveness through experiments.
\footnote{The article has been accepted by Frontiers of Computer Science (FCS), with the DOI: {10.1007/s11704-025-41167-w}.} \\
\end{abstract}


\begin{document}
\begin{sloppypar}

\section{Introduction}

Maintaining the diversity of a set of solutions is of great importance across a wide range of scenarios, such as reinforcement learning (RL)~\cite{NSLC,NSR-ES,DIAYN,dvd,NeuroevoSkillDiscovery,qsd,yao2023policy}, ensemble learning~\cite{ensemble-div1,ensemble-book,ensemble-datastream,ensemble-rl,margin}, data generation~\cite{DQD,crd-cgan-fcs,rainbow-teaming}, and combinatorial optimization~\cite{edo-tsp,nikfarjam2022computing}.
For example, diversity is a key requirement in open-ended learning~\cite{maley1999four,open-ended-artificial-evolution,towards-unifying}; in level generation tasks of RL, the generated levels should be diverse to improve the robustness of the trained policy~\cite{PLR,PLR2,qd4ge,dsage,accel,warehouse}; in ensemble learning, base learners should be diverse to make the ensemble learner perform well~\cite{ensemble-div0,ensemble-div1,DNS}; in safety fine-tuning of large language models, the adversarial prompts in the generated dataset should be diverse to improve the robustness against various style of attacks~\cite{rainbow-teaming}. Recently, many works leverage evolutionary algorithms~\cite{el-book} to enhance policy diversity in RL~\cite{moerl-scis,ccqd,refqd,erl-survey-lipy}.

Diversity optimization is a general and important problem of machine learning~\cite{ensemble-div1}. A lot of diversity optimization algorithms have been proposed to obtain a set of diverse solutions for various scenarios~\cite{qd-frontier,DIAYN,fu2023iteratively}. In general, most of the algorithms can be abstracted into three steps: 1) mapping the solutions (and the corresponding data collected during the evaluation of the solutions) into a latent feature space that can reflect the features of the solutions, 2) defining the diversity measure by combining the features and a distance (or similarity) metric, 3) obtaining a set of diverse solutions under the diversity measure. Among these steps, how to define the diversity measure is a longstanding problem~\cite{maley1999four,critical-factors,dvd,AURORA-CSC}. Since there are various methods for distance measure (e.g., $\ell_2$ distance and cosine similarity), the core of diversity definition lies in the definition of feature space. In this paper, we consider the application of diversity in RL and refer to the feature space as the behavior space following prior works~\cite{qd-frontier,cma-me,NeuroevoSkillDiscovery,refqd}.

There are currently two main categories of methods for defining behavior space.
One is to obtain the behavior space directly from the data themselves~\cite{DIAYN, AURORA-CSC}, i.e., totally data-driven approaches. However, without considering human requirements, the learned behavior spaces are usually far away from the ones required by humans. The other is to require an expert to define the behavior space directly~\cite{NSLC, ME, PGA-ME, DQD}. However, in many scenarios, even experts find it challenging to provide an appropriate behavioral metric, e.g., Real-Time Strategy (RTS) games like StarCraft~\cite{vinyals2017starcraft} and Multi-player Online Battle Arena (MOBA) games like Dota~\cite{berner2019dota}.

Despite the inability to accurately describe the behavior space within humans' minds and even provide an accurate definition, humans often have the ability to express a \emph{preference} based on their internal behavior space, i.e., they can tell which two solutions are more similar and which two are more diverse. For instance, in cooperative games between humans and AI, human players can provide feedback on which two agent partners exhibit similar behaviors (thus making human players feel bored and uninterested) and which two demonstrate diverse behaviors (thus creating a sense of excitement and intrigue).

In this paper, we propose the problem formulation of \emph{How to learn a behavior space from human feedback} for the first time. Then, we propose a general method Diversity from Human Feedback (DivHF) to solve the problem. DivHF learns the behavior space by querying the human preference and learning a model called behavior descriptor that is consistent with human preference. Then, we can combine the learned behavior descriptor with any distance measure to obtain the diversity measure, and use an arbitrary diversity optimization algorithm to obtain a set of diverse solutions based on it.

Our proposed method DivHF is general, which can cooperate with arbitrary diversity optimization algorithms. As an instantiation, we apply DivHF to Quality-Diversity (QD) optimization algorithms, which aim to find a diverse set of high-quality solutions of a problem and have many successful applications in RL. In particular, we conduct experiments with a representative QD algorithm MAP-Elites (ME)~\cite{ME,animal} on the popular QDax suite~\cite{accelerated-qd,Qdax}. The results demonstrate that the behavior space learned by DivHF is much more consistent with human requirements than the one learned directly from the data themselves without human feedback (i.e., Auto-encoder), and the final solutions obtained by the algorithm are more diverse under human preference. The effectiveness of DivHF is further verified by illustration of the learned behavior space and hyper-parameter sensitivity analysis.

Our contributions are as follows:
\begin{itemize}
    \item We formulate the problem of obtaining diversity from human feedback.
    \item We propose a general method, diversity from human feedback (DivHF), to solve the problem.
    \item Experimental results show that DivHF can learn the accurate behavior descriptor from human feedback and help the diversity optimization algorithms to obtain solutions that are diverse under human preference.
\end{itemize}

As a concurrent work with ours\footnote{This paper is a formal version of~\cite{divhf-arxiv}. Our work was first published on arXiv on Oct 10, 2023~\cite{divhf-arxiv}, while QDHF was done on Oct 18, 2023~\cite{qdhf-arxiv}.}, QDHF~\cite{qdhf} also leverages human feedback to derive diversity metrics. Compared with our work, QDHF uses another loss function instead, which can be seen as a clipped variant of our simple loss function in Eq.~(\ref{eq:Vanilla-loss}).
We compare it with our method in the experiments and show that ours performs better.

\section{Preliminaries}
We focus on the diversity optimization algorithms, which can be abstracted into two parts: mapping the solutions $\{\sol_i\}_{i=1}^N$ (and the information $\{\data_i\}_{i=1}^N$ collected during the evaluation of the solutions on the same set of data) into behavior vectors $\{\bd_{\param}(\sol_i, \data_i)\}_{i=1}^N$ by a behavior descriptor $\bd_{\param}(\cdot, \cdot)$ parameterized by $\param$,
and then optimizing the diversity $Div(\{\bd_{\param}(\sol_i, \data_i)\}_{i=1}^N)$ in the behavior space. For convenience, we use $\x$ to denote the pair $(\sol, \data)$ of a solution $\sol$ and its corresponding evaluation information $\data$.

Considering the human requirements, for two pairs of solutions $(\x_1, \x_2)$ and $(\x_3, \x_4)$, we denote $(\x_1, \x_2) \succ (\x_3, \x_4)$ if and only if $(\x_1, \x_2)$ is more similar than $(\x_3, \x_4)$ under the human preference.

\section{Related Work}

\subsection{Diversity Optimization}
Most diversity optimization algorithms can be viewed as optimizing the diversity metrics in a latent space, e.g., behavior space. The mapping from the solutions to the behavior is the common process of the algorithms, which is critical and determines whether the algorithms can generate a set of diverse solutions that meet human requirements.

Take a popular family of diversity optimization algorithms, i.e., QD algorithms~\cite{ME,animal,NSLC, qd-survey-optimization} as examples, which aim to generate a set of high-quality and diverse solutions. Given a fitness function $f$ to be maximized and an expert-defined behavior descriptor $\bd$, the goal of QD is to generate a set of solutions that cover the space of behavior descriptors and have high fitness values.

ME~\cite{ME,animal}, the most well-known QD algorithm, maintains an archive by discretizing the behavior space into $M$ cells $\{\bdspace_i\}_{i=1}^{M}$ and storing at most one solution in each cell. ME aims to fill the cells with high-quality solutions. Thus, the goal of ME is to maximize the QD-Score $\sum\nolimits_{i=1}^M f(\sol_i)$, where $\sol_i$ represents the solution contained within the cell $\bdspace_i$, i.e., $\bd(\sol_i) \in \bdspace_i$. If a cell $\bdspace_i$ does not contain a solution $\sol_i$, then $f(\sol_i)$ is considered as $0$. For simplicity, the fitness value $f(\cdot)$ is assumed (or converted) to be non-negative to prevent the solutions from decreasing the QD-Score. As a type of evolutionary algorithm, the main process of ME is to select parent solutions from the archive, generate offspring solutions through variation operators, evaluate the offspring solutions, and update the archive.

There are many works focusing on improving the sample efficiency of QD by refining the parent selection~\cite{qd-survey-framework,EDO-CS,NSS}, variation operators~\cite{ME-ES,cma-me,PGA-ME,DQD,qd-pg}, or by applying cooperative coevolution~\cite{ccqd}. Recently, Qian et al.~\cite{qd-theory-ijcai24} theoretically studied the benefit of QD on optimization for the first time, and Wang et al.~\cite{refqd} tried to improve the resource efficiency of QD. However, all the works require a behavior space pre-defined by experts.

\subsection{Behavior Space of Diversity Optimization}
According to the way the behavior space is defined, the algorithms can be usually categorized into two types. 
The first type is to obtain the behavior space directly from the data themselves. For example, AURORA~\cite{AURORA,AURORA-CSC} uses auto-encoder to learn the behavior space in an unsupervised manner from raw sensory data; 
RUDA~\cite{RUDA} considers the relevance of the solutions in downstream tasks, varying the distance metric to generate more relevant solutions;
Many skill discovery methods (e.g., DIAYN~\cite{DIAYN} and DADS~\cite{dads}) sample several diverse behaviors from a behavior space, and train a behavior-conditioned policy to maximize the mutual information between the behaviors and the trajectories of the policy.
However, as these types of methods do not consider the human requirements of diversity, the obtained diversity is usually far away from the one required by human, making them less useful in many practical and complex scenarios.

The second type considers the requirements of humans by requiring an expert to define the behavior space directly. For example, given an expert-defined behavior space, ME~\cite{ME,animal} and NSLC~\cite{NSLC} find a set of diverse solutions that cover the behavior space. In robotic training, in order to train a set of policies that help recover quickly from damage, the experts use the touchdown time of feet as the behavior descriptor and use ME to obtain a set of solutions with varying frequency of feet usage~\cite{animal,PGA-ME,NeuroevoSkillDiscovery,ccqd}. Interactive Constrained ME~\cite{ICME-CoG,ICME-TG} enables the human designer to flexibly change the feature dimensions (i.e., behavior space) for content generation.
However, in many scenarios, it is hard to define a proper behavior space, even for the experts. 

\subsection{Learning from Human Feedback}
Preference-based RL is proposed to provide human-preferred objectives to RL agents~\cite{pbrl, DRLHP, open-problems}. In this approach,  humans are asked to provide their preferences regarding pairs of agents' historical trajectories. By using human feedback, a reward model is learned and utilized to provide learning signals to the agents. Preference-based RL offers an effective means to learn from human intentions, rather than relying solely on explicitly designed rewards. Its effectiveness has been demonstrated in various domains, such as robotics control~\cite{solar} and chatbots~\cite{ouyang, instructgpt}. Notably, reinforcement learning from human feedback (RLHF) has emerged as a key strategy for fine-tuning large language models~\cite{gpt4, gemini}.

In this work, we investigate the important question for the first time: How to define a proper behavior space for diversity optimization? Note that defining a proper behavior space directly is difficult even for an expert in many scenarios. Inspired by RLHF, we propose the DivHF method to solve this problem by learning the behavior space and optimizing the diversity from human feedback. Unlike RLHF, DivHF considers finding the diversity measure rather than a reward function (or reward bonus~\cite{CDP}). DivHF asks the human to determine which \emph{pair of policies} is more diverse, rather than which policy is preferred. DivHF predicts the behaviors of solutions to be used by diversity optimization algorithms, rather than the rewards of state-action pairs. 

\section{DivHF}

\subsection{Problem Formulation}

\begin{figure*}[htb]
    \centering
    \includegraphics[width=0.6\linewidth]{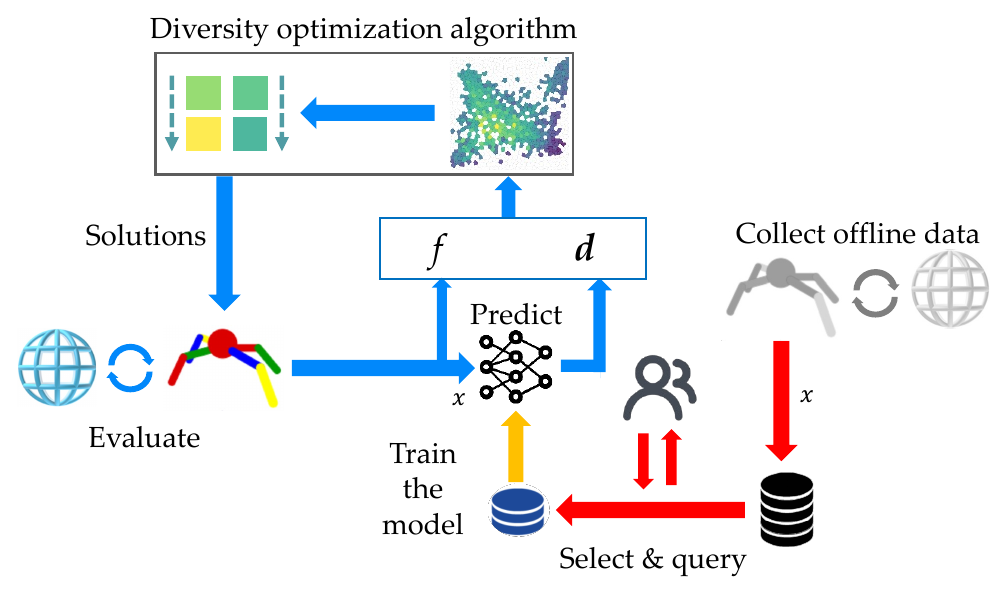} 
    \caption{The workflow of DivHF.}
    \label{fig:workflow-main}
\end{figure*}

In this paper, we explore the following problem: \emph{How to obtain a set of diverse solutions based on human preference?} This problem is common and natural since humans often find it challenging to define a proper behavior space or assign exact behavior values. Instead, it is much easier for humans to distinguish which two solutions are the most similar and which two are the most diverse from a set of solutions, which is expressed through \emph{human feedback}. This feedback reflects their preference for the \emph{diversity}. If we can learn a behavior descriptor to establish a behavior space, we can easily define similarity (or distance) using various measures such as cosine similarity or Euclidean distance, and obtain a set of diverse solutions based on the learned behavior descriptor by any diversity optimization algorithms. Therefore, our main objective is to efficiently address the problem of \emph{How to learn a behavior descriptor from human feedback?}.

To be more clear, given the data $\{\x_i\}_{i=1}^N$, we ask for the human preference. If the human thinks $(\x_1, \x_2)$ is more similar than $(\x_3, \x_4)$, then we want the similarity between $\bd_\param (\x_1)$ and $\bd_\param (\x_2)$ to be larger than that between $\bd_\param (\x_3)$ and $\bd_\param (\x_4)$, to be consistent with the human preference.

\subsection{Overall Method}

We propose the DivHF method to learn the behavior descriptor and optimize the diversity from human feedback. The main idea is to select some data to query for human preference and train a model of behavior descriptor that is consistent with human preference.
As shown in Figure~\ref{fig:workflow-main}, the main process of DivHF can be summarized as follows.
\begin{enumerate}
    \item Select some solutions (along with the corresponding evaluation data, e.g., trajectories) as examples to query human preference (red arrows).
    \item Train a behavior descriptor model to extract the behaviors of solutions according to the human preference data (yellow arrows).
    \item Optimize the solutions based on the learned behavior descriptor by using an arbitrary diversity optimization algorithm (blue arrows).
\end{enumerate}

\subsection{Data Collection: Query the Human Preference}

Considering the requirements of diversity optimization algorithms, a behavior descriptor model takes the solutions and/or the evaluation data as input features, and the behavior predictions as outputs. The model should be designed according to the structure of features. For example, if the evaluation data are trajectories of reinforcement learning, we can use bidirectional stacked Long Short-Term Memory (LSTM)~\cite{LSTM, stacked-lstm} as the model.

To train the model, we need to select some solutions and their evaluation data to query human preference. DivHF takes a set of three solutions $(\x_1, \x_2, \x_3)$ as a query, and lets a human determine which two are the most similar and which two are the most diverse. If we take only two solutions as a query, the human will be required to tell their exact similarity value, which is hard for the human. If we take too many solutions as a query, it will take more time for the human to determine.

\subsection{Training the Behavior Descriptor}
\label{subsec:divhf-training}

\begin{figure*}[tb]
    \centering
    \includegraphics[width=0.94\linewidth]{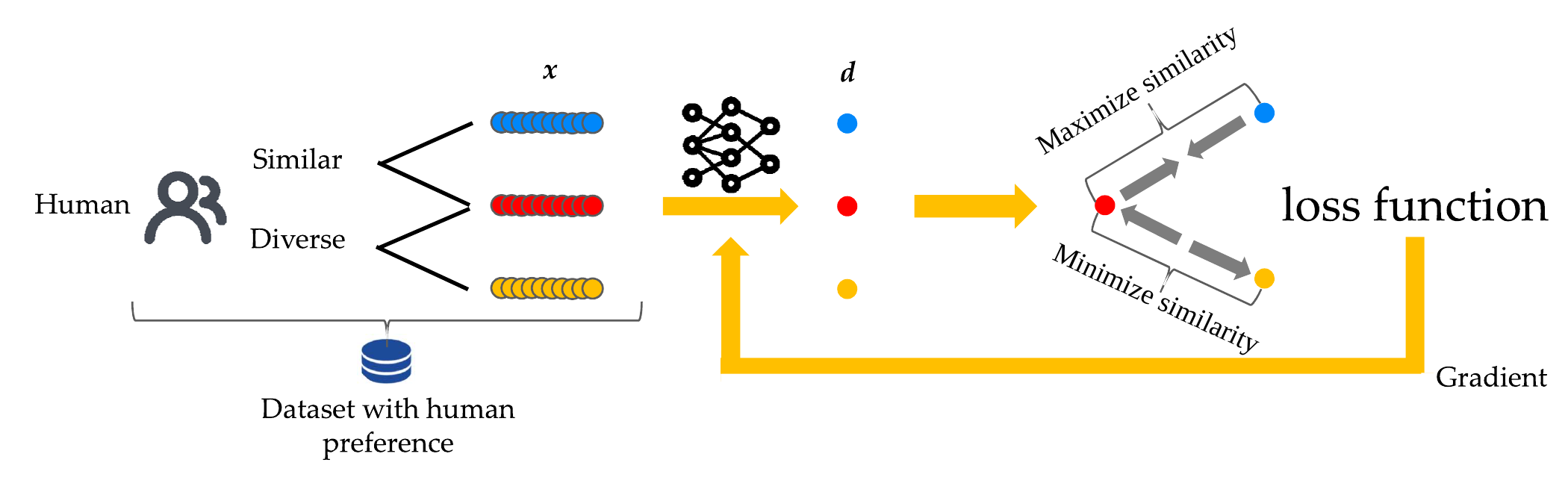}
    \caption{The training process of DivHF.}
    \label{fig:workflow-training}
\end{figure*}

With the human preference data, we want to train the model that can give behaviors that are consistent with human preference by a loss function and an optimizer. As shown in Figure~\ref{fig:workflow-training}, given a piece of human preference, we want to maximize the similarity of the behaviors of the most similar solutions and minimize the similarity of the behaviors of the most diverse ones according to the human preference. Given a similarity metric $\simfn(\cdot, \cdot)$, we can predict the \emph{pair-wise similarity} $\simfn (\bd_\param(\x_i), \bd_\param(\x_j))$. Assuming, without loss of generality, that $(\x_1, \x_2)$ is the most similar and $(\x_1, \x_3)$ is the most diverse, then $\simfn(\bd_\param(\x_1), \bd_\param(\x_2))$ and $\simfn(\bd_\param(\x_1), \bd_\param(\x_3)))$ should be maximized and minimized, respectively. Thus, a simple loss function is to optimize the corresponding similarity metric simultaneously:
\begin{equation}
    \mathcal L (\param) = \sum_{(\x_1, \x_2, \x_3) \in \mathcal D} \simfn(\bd_\param(\x_1), \bd_\param(\x_3)) - \simfn(\bd_\param(\x_1), \bd_\param(\x_2)).
    \label{eq:Vanilla-loss}
\end{equation}
However, since the preferences given by humans are relative, optimizing the proportion is better, which can eliminate the effects of scaling.

Another way is to use the Bradley-Terry model~\cite{btmodel}, which is similar to prior works in RLHF~\cite{DRLHP, rhp, pebble} and usually called InfoNCE loss~\cite{InfoNCE}. For a piece of preference data $(\x_1, \x_2, \x_3)$, we can calculate the predicted behavior $\{\bd_\param(\x_i)\}_{i=1}^3$.
Then, we can predict the similarities $\{\simfn (\bd_\param(\x_i), \bd_\param(\x_j))\}_{i,j \in \{1, 2, 3\}, i \neq j}$ of the pairs $\{(\x_i, \x_j)\}_{i,j \in \{1, 2, 3\}, i \neq j}$. We can consider $\simfn (\bd_\param(\x_i), \bd_\param(\x_j))$ as the underlying factor that reflects the human preference between $(\x_i, \x_j)$ and other pairs. We assume that the human preferences are influenced by the power of these factors. Consequently, we can predict the probability of preferences by applying the soft-max function to the respective factors:
\begin{align*}
    &\Prob_\param [(\x_i, \x_j) \succ (\x_i, \x_k)] = \\
    &\frac{
        \exp (\lambda \simfn (\bd_\param(\x_i), \bd_\param(\x_j)))
    }{
        \exp (\lambda \simfn (\bd_\param(\x_i), \bd_\param(\x_j))) + \exp (\lambda \simfn (\bd_\param(\x_i), \bd_\param(\x_k)))
    },
\end{align*}
where $(\x_i, \x_j) \succ (\x_i, \x_k)$ denotes that $(\x_i, \x_j)$ is more similar than $(\x_i, \x_k)$, and $\lambda$ is the temperature that adjusts the discrimination between positive and negative samples.
The goal is to learn $\bd_\param$ that minimizes the cross-entropy between the predictions and the human preferences: 
\begin{equation}
    \mathcal L (\param) = - \sum_{(\x_1, \x_2, \x_3) \in \mathcal D} \log \Prob_\param [(\x_1, \x_2) \succ (\x_1, \x_3)],
    \label{eq:DivHF-loss}
\end{equation}
where we assume, without loss of generality, that $(\x_1, \x_2)$ is the most similar and $(\x_1, \x_3)$ is the most diverse. Note that we only use the most similar pair and the most diverse pair, as they are easier to determine by humans.

\section{Experiments}
To examine the performance of DivHF, we conduct experiments on the popular QDax domain~\cite{accelerated-qd, Qdax}, including HalfCheetah, Walker2D, Ant, and Humanoid. The goal of the tasks is to generate a set of robotic policies that are diverse in the frequency of the usage of each foot, in order to enable the robot to recover quickly from damage. We collect 25k triples of trajectories offline to train the behavior descriptor model. For convenience, we define an oracle behavior space for each task, i.e., the fraction of time each foot touches the ground. We also use a synthetic oracle whose preferences exactly reflect the similarity of the oracle behaviors. That is, when the method queries for a preference, we immediately reply which two trajectories have the most similar oracle behaviors and which two have the most diverse ones.

In our experiments, we use the cosine similarity as the similarity metric. We use a bidirectional two-layer-stacked LSTM~\cite{LSTM, stacked-lstm} followed by a two-layer MLP to extract the feature from the temporal trajectory data, where the feature sizes of LSTM layers are both $512$, and the hidden layer size of MLP layers is $32$. The network structure is:
\begin{align*}
    \text{trajectory} &\to \textbf{LSTM}(512, 512) \to \textbf{MLP}(32) \to \textbf{tanh} \\
    &\to \textbf{MLP}(\text{dimension of behavior}) \to \text{behavior}
\end{align*}
The dimension of the learned space is the same as the oracle behavior space. All the experiments are conducted on an NVIDIA RTX A6000 GPU with an AMD EPYC 7513 CPU.

To evaluate the effectiveness of DivHF, we compare the following methods. \textbf{DivHF}: The proposed method that uses cross-entropy in Eq.~(\ref{eq:DivHF-loss}) as the loss function. \textbf{DivHF w/o CE}: The same as DivHF, except that it uses simple loss in Eq.~(\ref{eq:Vanilla-loss}) as the loss function. \textbf{DivHF w/o BT}: The same as DivHF, except that it uses simple LSTM as the model. \textbf{DivHF-Vanilla}: The same as DivHF, except that it uses simple loss in Eq.~(\ref{eq:Vanilla-loss}) as the loss function and simple LSTM as the model. \textbf{Auto-encoder}: The method that concatenates a decoder after the behavior descriptor model and learns the behavior space directly by self-supervision without human preference. \textbf{Oracle}: The oracle behavior descriptor itself. We also compare the concurrent work QDHF~\cite{qdhf}, which is the same as DivHF, except that it uses a clipped variant of our simple loss function in Eq.~(\ref{eq:Vanilla-loss}) instead:
\begin{align*}
    \mathcal L (\param) = \sum_{(\x_1, \x_2, \x_3) \in \mathcal D} \max\big\{0, m + &\simfn(\bd_\param(\x_1), \bd_\param(\x_3))\\
     - &\simfn(\bd_\param(\x_1), \bd_\param(\x_2))\big\}.
\end{align*}
where $m$ is a constant bound.

After training, all the behavior description methods are in conjunction with ME~\cite{ME, animal} to generate a set of diverse solutions. Specifically, we use the learned mapping instead of the expert-defined behavior mapping to obtain the behaviors of the solutions in the ME framework.

The experiments aim to answer the following two Research Questions (RQs): 1) Whether the learned behavior descriptors are consistent with human preferences? 2) When applied to diversity optimization algorithms (e.g., Quality-Diversity algorithms), whether the solutions are diverse under human preference?

\subsection{RQ1: Whether the Learned Behavior Descriptors are Consistent with Human Preferences?}

We first train the models with the human preference data and use the following metrics to evaluate whether the behavior descriptors learned by different methods are consistent with human preference.
\begin{itemize}
    \item \textbf{Most Similar Accuracy}: The accuracy of selecting the most similar pair in each triple of data, which is calculated by the proportion of making correct selections. 
    \item \textbf{Most Diverse Accuracy}: The accuracy of selecting the most diverse pair in each triple of data. 
    \item \textbf{Preference Accuracy}: The accuracy of selecting the most similar pair and the most diverse pair in each triple of data at a time. Note that as there are three pairs in each triple of data, selecting the most similar pair and the most diverse pair correctly also implies predicting the preference of three pairs correctly. 
    \item \textbf{Pair-wise Accuracy}: The accuracy of predicting the preference of two pairs, i.e., whether one pair is more diverse than the other under human preference, in each triple of data.
\end{itemize}

\begin{table*}[htbp]
\centering
\caption{Preference Accuracy comparisons (mean) of different methods on the test sets under different environments. \textbf{Bold} texts denote the best method. Average Ranking is the average Preference Accuracy ranking across all environments. Average Ratio is the average Preference Accuracy ratio between DivHF and other methods across all environments.}
\begin{tabular}{c|cccccc}
\toprule
Environment & Auto-encoder & QDHF & DivHF-Vanilla & DivHF w/o CE & DivHF w/o BT & DivHF \\ \midrule
HalfCheetah Uni & $0.171$ & $0.356$ & $0.347$ & $0.487$ & $0.265$ & $\textbf{0.563}$ \\
Walker2D Uni    & $0.246$ & $0.558$ & $0.501$ & $0.521$ & $0.594$ & $\textbf{0.655}$ \\
Ant Uni         & $0.365$ & $0.612$ & $0.453$ & $0.444$ & $0.644$ & $\textbf{0.719}$ \\
Humanoid Uni    & $0.251$ & $0.310$ & $0.288$ & $\textbf{0.395}$ & $0.261$ & $0.312$ \\
\midrule
Average Ranking & $6.00$ & $3.00$ & $4.25$ & $3.00$ & $3.50$ & $\textbf{1.25}$ \\
Average Ratio & $0.498$ & $0.832$ & $0.733$ & $0.886$ & $0.777$ & $\textbf{1.000}$ \\ \bottomrule
\end{tabular}
\label{tab:accuracy}
\end{table*}

\begin{figure*}[h!]
    \centering    
    \includegraphics[width=0.96\linewidth]{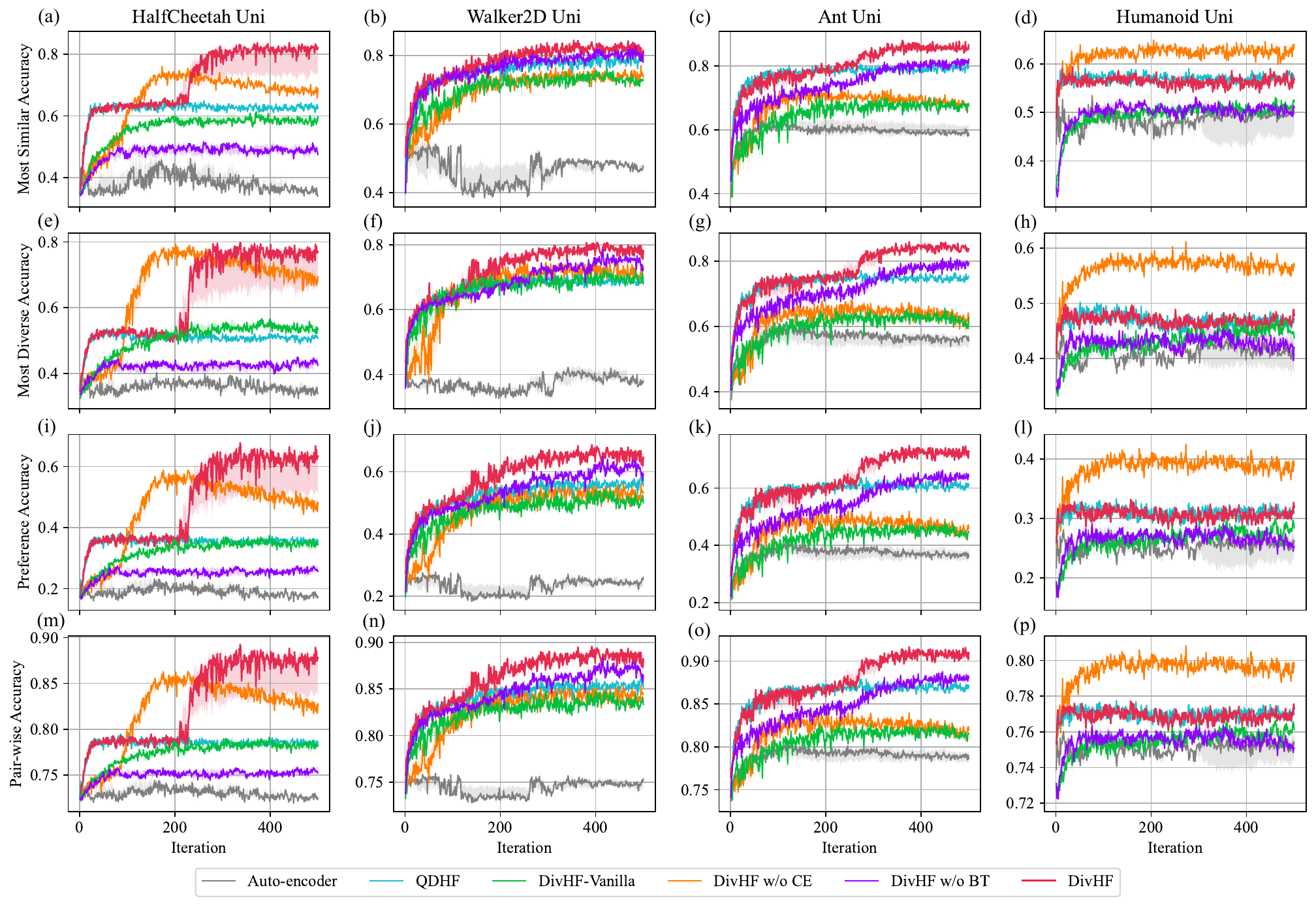}
    \caption{Most Similar Accuracy (a--d), Most Diverse Accuracy (e--h), Preference Accuracy (i--l), and Pair-wise Accuracy (m--p) of different methods on the test sets under different environments. The medians and the first and third quartile intervals are depicted with curves and shaded areas, respectively.}
    \label{fig:accuracy}
\end{figure*}

The results are shown in Figure~\ref{fig:accuracy}. The accuracy of all variants of DivHF is better than Auto-encoder in all environments, indicating that DivHF can really learn from human preferences and predict behaviors that are consistent with human preferences. DivHF achieves the best accuracy in most cases, except the environment Humanoid Uni where DivHF w/o CE (i.e., DivHF using simple loss in Eq.~(\ref{eq:Vanilla-loss})) performs better.
This may be due to the fact that it has the largest search space and has a relatively different landscape.
Note that DivHF shows sudden boost in performance in HalfCheetah and Ant, which may be because that LSTM networks are difficult to train and behavior spaces are difficult to learn. The sudden boost in training LSTM networks can also be observed in other works~\cite{lstm-pomdp2,lstm-pomdp3}.
DivHF also performs better than QDHF among these four environments, showing that optimizing the proportion is better than optimizing the difference. Overall, DivHF has the best average ranking on Preference Accuracy, as shown in Table~\ref{tab:accuracy}.

\begin{table*}[htb]
\centering
\caption{QD-Score comparisons (mean) of different methods on the test sets under different environments. \textbf{Bold} texts denote the best method except for Oracle. Average Ranking is the average QD-Score ranking across all environments. Average Ratio is the average QD-Score ratio between DivHF and other methods across all environments.}
\begin{tabular}{c|cccccc|c}
\toprule
Environment & Auto-encoder & QDHF & DivHF-Vanilla & DivHF w/o CE & DivHF w/o BT & DivHF      & Oracle      \\ \midrule
HalfCheetah Uni & $749,067$   & $783,708$ & $\textbf{1,368,209}$   & $946,628$   & $1,070,816$  & $852,532$ & $2,550,238$ \\
Walker2D Uni    & $209,583$   & $\textbf{371,790}$ & $227,111$    & $324,036$   & $258,224$   & $353,520$ & $700,442$  \\
Ant Uni         & $26,170$    & $266,309$ & $61,090$     & $61,967$    & $247,828$   & $\textbf{278,938}$ & $796,666$  \\
Humanoid Uni    & $289,382$   & $212,199$ & $313,680$    & $\textbf{363,041}$   & $334,443$   & $349,922$ & $725,198$  \\ \midrule
Average Ranking & $6.75$ & $4.50$ & $4.75$ & $3.75$ & $4.00$ & $\textbf{3.25}$     & $1.00$      \\
Average Ratio   & $0.598$ & $0.883$ & $0.841$ & $0.821$ & $0.958$ & $\textbf{1.000}$     & $2.475$      \\ \bottomrule
\end{tabular}
\label{tab:algo-metrics}
\end{table*}

\subsection{RQ2: When Applied to Classical Algorithms, whether the Solutions are Diverse under Human Preference?}

Next, we use the learned behavior models as the behavior descriptors of classical QD algorithms and examine whether they can obtain a set of solutions that are diverse under human preference. In order to examine the diversity, after each generation, we map the solutions to the oracle behavior space and put them into an oracle CVT archive. Then, we calculate the classical QD metrics to show the diversity of the solutions under the human preference.

The following QD metrics are considered. 1) \textbf{QD-Score}: The total sum of the fitness values across all solutions in the oracle CVT archive. It can measure both the quality and the diversity of the solutions. 2) \textbf{Max Fitness}: The largest fitness value of the solutions in the archive. It can measure the quality of the solutions. 3) \textbf{Coverage}: The total number of solutions in the oracle CVT archive. It can measure the diversity of the solutions.

\begin{figure*}[htb]
\centering    
\includegraphics[width=0.95\linewidth]{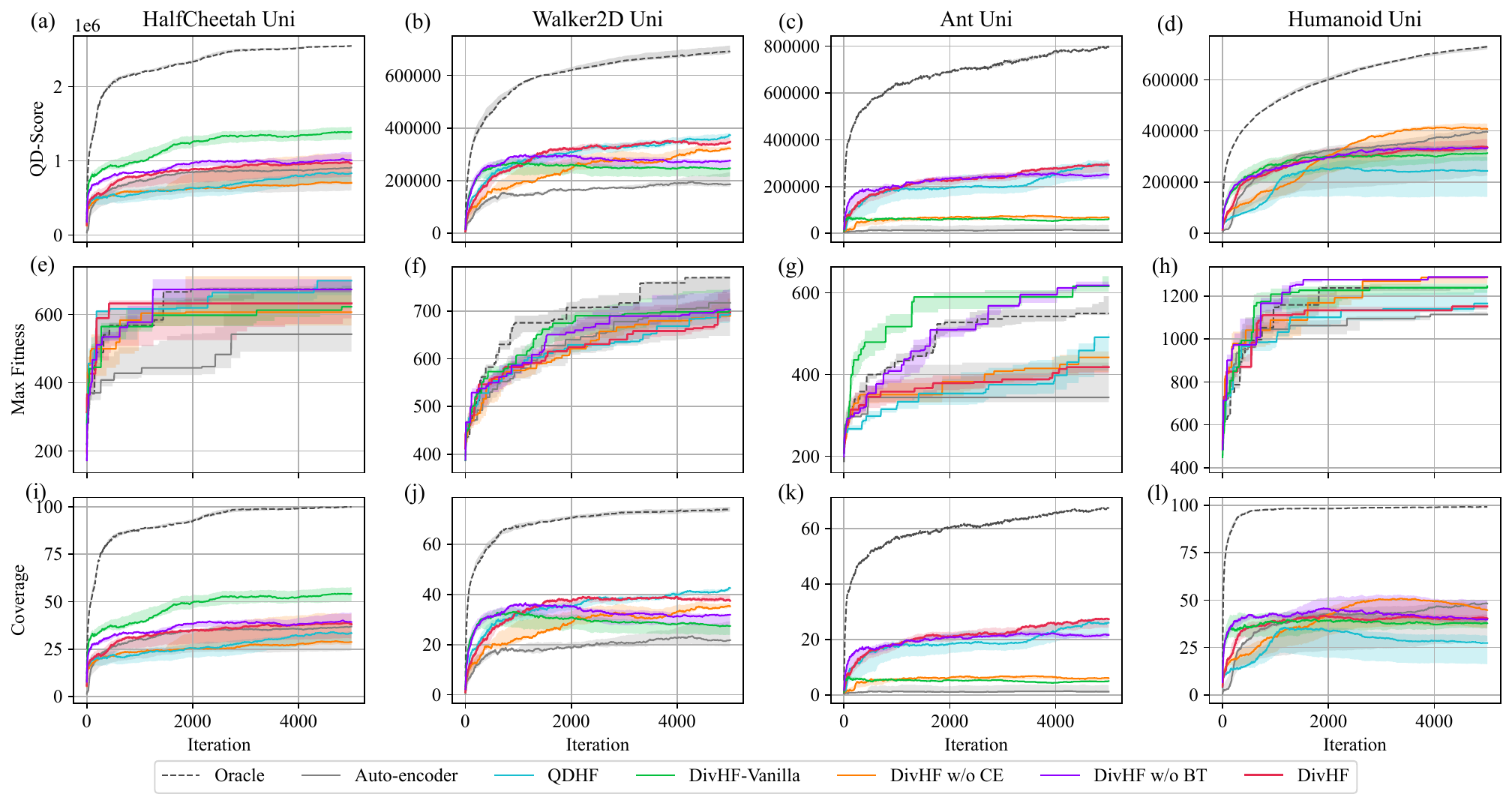}
\caption{Performance comparisons in terms of QD-Score (a--d), Coverage (e--h), and Max Fitness (i--l) under the four environments. The medians and the first and third quartile intervals are depicted with curves and shaded areas, respectively.}
\label{fig:algo-metrics}
\end{figure*}

The results are shown in Figure~\ref{fig:algo-metrics}. The QD-Score and Coverage are more related to the goodness of the learned behavior. In general, the higher the accuracy of the learned behavior, the higher QD-Score obtained. As shown in Tables~\ref{tab:accuracy} and \ref{tab:algo-metrics}, the average rankings of the Preference Accuracy of behavior and QD-Score are similar: Oracle $<$ DivHF $<$ QDHF, DivHF w/o CE, DivHF w/o BT $<$ DivHF-Vanilla $<$ Auto-encoder.
The rankings of the average ratios are also similar, except for DivHF w/o CE, which has a worse average ratio due to its poor performance in Ant.
We can also observe from Figure~\ref{fig:algo-metrics} that compared with the oracle behavior descriptor, the QD-Score and Coverage of DivHF still have a large gap, implying that further improvement is expected in the future.

\begin{figure*}[htb]
    \centering
    \includegraphics[width=0.95\linewidth]{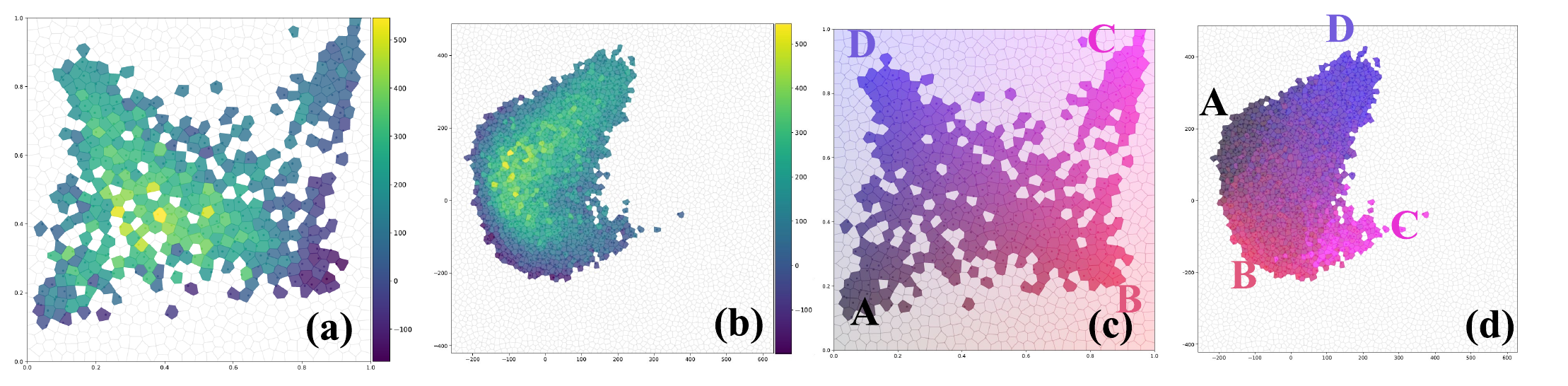}
    \vspace{-0.1em}
    \caption{Visualization of the solutions obtained from DivHF. The sub-figures~(a) and (c) denote the oracle behavior space, and the sub-figures~(b) and (d) denote the learned behavior space. The color in sub-figures~(a) and (b) denote the quality of the solutions. We put gradient colors on sub-figure~(c), and map the solutions with their colors to sub-figure~(d).}
    \label{fig:solution-visualization}
\end{figure*}

Furthermore, we visualize the solutions in the oracle behavior space and the learned behavior space. As shown in Figure~\ref{fig:solution-visualization}, the sub-figures~(a) and (c) denote the oracle behavior space, and the sub-figures~(b) and (d) denote the learned behavior space. The color in sub-figures~(a) and (b) denote the quality of the solutions. In sub-figure~(c), we use different colors (black, red, pink, and purple) to color the oracle behavior space. Then, we map the solutions with their colors to the learned behavior space, i.e., sub-figure~(d). Sub-figures~(c) and (d) show a clear resemblance between the oracle space and the learned behavior space: the oracle behavior space can be recovered roughly by rotating the learned behavior space counterclockwise. This indicates that the learned behavior space captures the essence of the oracle behavior space.

\subsection{Additional Studies}

\begin{figure*}[h!]
    \centering    \includegraphics[width=0.95\linewidth]{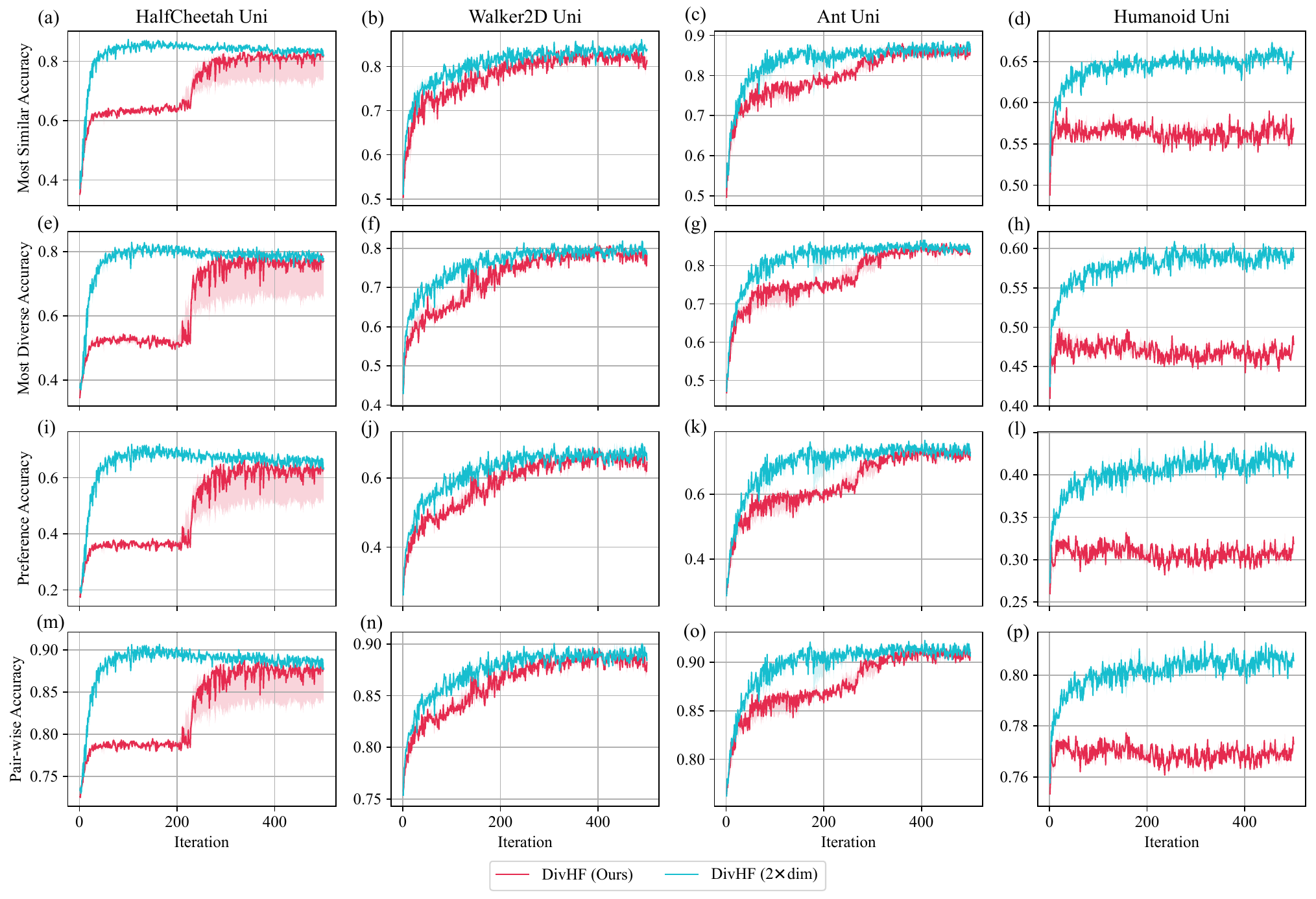}
    \vspace{-0.1em}
    \caption{Most Similar Accuracy (a--d), Most Diverse Accuracy (e--h), Preference Accuracy (i--l), and Pair-wise Accuracy (m--p) of DivHF with different dimensions of behavior descriptor, where DivHF and DivHF (2 $\times$ dim) denote that the dimension of the learned behavior descriptor is the same as and twice of the dimension of the oracle behavior descriptor, respectively. The medians and the first and third quartile intervals are depicted with curves and shaded areas, respectively.}
    \label{fig:bd-len}
\end{figure*}

\begin{figure*}[h!]
    \centering    
    \includegraphics[width=0.95\linewidth]{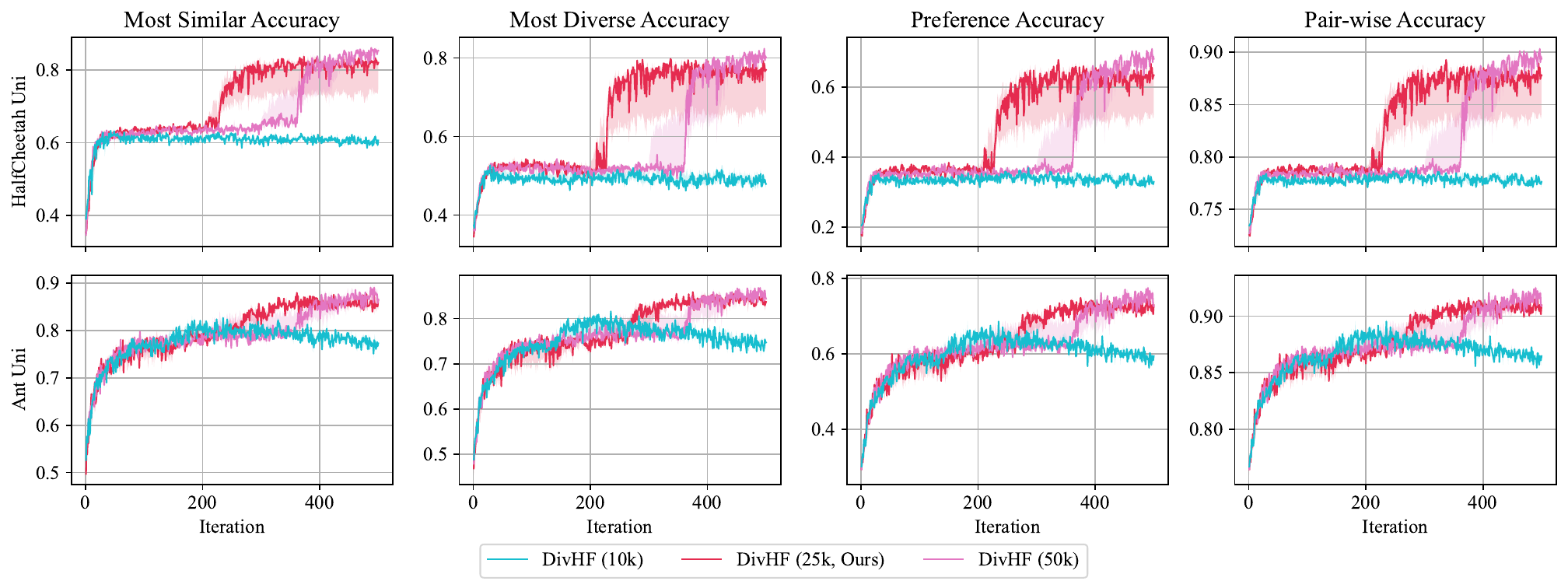}
    \caption{Accuracy of DivHF using different sizes of training set, i.e., different number of queries in training. The medians and the first and third quartile intervals are depicted with curves and shaded areas, respectively. 25k is used in our experiments.}
    \label{fig:n-queries}
\end{figure*}

\textbf{Dimension of Behavior Space.}
We examine the influence of the dimension of the behavior space. We compare the setting of using the same dimension as the oracle behavior and that of using a doubled dimension, which are denoted as DivHF and DivHF (2$\times$dim), respectively. Figure~\ref{fig:bd-len} shows that a higher dimension can lead to better accuracy due to the better representation ability of the model. However, a behavior space with too high dimension may be hard for humans to understand, and also hard for QD algorithms to optimize.

\textbf{Number of Queries.}
We further examine the influence of the number of queries used for training. We compare the settings that use 10k, 25k, and 50k queries. As expected, Figure~\ref{fig:n-queries} shows that using a too small number of queries may lead to poor performance. The use of many human queries is one of the limitations of the proposed method, and we will consider reducing the number of human queries in future work.

\section{Discussion}

Diversity plays an important role in many scenarios and usually relies on a well-defined behavior descriptor, which is, however, challenging in practice. In this paper, we introduce a natural way of defining diversity from human feedback for the first time. We propose a general method called DivHF, which selects triples of data to query humans about which two are the most similar and which two are the most diverse, and trains the behavior descriptor with simple linear loss or cross-entropy loss to make it consistent with human preference. DivHF can cooperate with arbitrary diversity optimization algorithms, and we apply it to QD algorithms as an instantiation. Experimental results on QDax show that DivHF can learn the accurate behavior descriptor from human feedback, and performs better than the concurrent work QDHF.

This paper is a preliminary work that obtains diversity from human feedback. There are many interesting future works, including improving the accuracy of the models, reducing the number of human queries, combining different loss functions to leverage their advantages together, applying to other diversity optimization algorithms and other scenarios, and handling the diverse underlying preferences of different humans effectively.

\begin{acknowledgement}
This work was supported by the National Science Foundation of China (62276124, 62272210), and the Fundamental Research Funds for the Central Universities (14380020). Ke Xue was supported by the National Science Foundation of China for PhD Students (624B2069) and Young Elite Scientists Sponsorship Program by CAST for PhD Students. Chao Qian is the corresponding author.
\end{acknowledgement}

\bibliographystyle{fcs}
\bibliography{fcs}

\begin{biography}{wangrj} Ren-Jian Wang received the B.Sc. degree in Computer Science and Technology from School of Computer Science and Technology, Dalian University of Technology in 2022. He is currently pursuing the Ph.D. degree at the School of Artificial Intelligence, Nanjing University. His current research interests mainly include evolutionary algorithms and quality diversity optimization.
\end{biography}

\begin{biography}{xuek} Ke Xue received the B.Sc. degree in Mathematics and Applied Mathematics from School of Mathematics, Sun Yat-Sen University in 2019. He is currently pursuing the Ph.D. degree with the School of Artificial Intelligence, Nanjing University, Nanjing, China. His current research interests mainly include machine learning and black-box optimization.
\end{biography}

\begin{biography}{wangyt} Yutong Wang received her BSc degree from the School of Mathematical Sciences, University of Science and Technology of China in 2020 and her MSc degree from the School of Artificial Intelligence, Nanjing University in 2023. She is currently working as an engineer at Huawei Noah’s Ark Lab, Shenzhen, China. Her research mainly interests include machine learning, reinforcement learning and evolutionary algorithms.
\end{biography}

\begin{biography}{yangp} Peng Yang is a tenure-track assistant professor at Southern University of Science and Technology (SUSTech), China. He received his B.Sc. and Ph.D. degrees from University of Science and Technology of China in 2012 and 2017, respectively. His research interests include Quantitative Finance and Multi-agent Simulation. He has published over 40 papers in top journals and conferences. He has served as the reviewer for top-tier journals and the PC member of top conferences (NeurIPS, ICLR, and ICML). He is the vice chair of IEEE CIS Evolutionary Learning Task Force, an executive member of CCF Computational Economics Committee and CCF AI Multi-agent System Committee, and a member of IEEE CIS Evolutionary Computation Technical Committee. He is an IEEE Senior Member.
\end{biography}

\begin{biography}{haobofu} Haobo Fu received the PhD degree in computer science from University of Birmingham, Birmingham, U.K., in 2014. From 2014 to 2016, he was a senior researcher with the Big Data
Laboratory, Baidu, Inc., Beijing, China. Since 2016, he has been working in the AI Platform Department, Tencent Inc., Shenzhen, China. He is now a tech lead and a principal researcher. His main interests are new innovations in AI for the game industry. These include innovative game AI applications and related research topics, such as machine learning, reinforcement learning, game theory, evolutionary computation, etc.  He has authored or co-authored over 40 refereed publications.
\end{biography}

\begin{biography}{qiangfu} Qiang Fu is currently the Director of Game AI center, Tencent AI Lab. His research interest includes reinforcement learning, domain data mining and multiagent systems. Leon's current focus is leading the game AI R\&D team to study intelligent AI in games and corresponding applications using deep learning, reinforcement learning and game theory.
\end{biography}

\begin{biography}{qianc} Chao Qian is a Professor in the School of Artificial Intelligence, Nanjing University, China. His research mainly focuses on building the theoretical foundation of evolutionary learning, and application to solve complex black-box optimization problems in industry and science. He has published one book “Evolutionary Learning: Advances in Theories and Algorithms”, and over 60 first/corresponding-authored papers in top-tier journals (PNAS, AIJ, ECJ, TEC, Algorithmica, TCS) and conferences (AAAI, IJCAI, ICML, NeurIPS, ICLR, DAC). He has won the ACM GECCO’11 Best Theory Paper Award, and the works on chip placement have won the DATE’25 (one leading EDA conference) Best Paper Award and the 21st ACM SIGEVO Humies Bronze Award. He serves on the editorial board of Artificial Intelligence Journal, Evolutionary Computation Journal, IEEE Trans. Evolutionary Computation, IEEE Computational Intelligence Magazine, etc. He is the founding chair of IEEE Computational Intelligence Society (CIS) Task Force on Evolutionary Learning, and was also the chair of IEEE CIS Task Force on Theoretical Foundations of Bio-inspired Computation. He has regularly given tutorials and co-chaired special sessions at CEC, GECCO and PPSN, given an Early Career Spotlight Talk at IJCAI’22, and will be a Program Co-Chair of PRICAI’25. He has helped solve complex optimization problems (e.g., supply chain, wireless network, and chip register optimization) in Huawei, and won Huawei Spark Award twice. He is a recipient of the NSFC for Excellent Young Scholars (2020) and CCF-IEEE CS Young Computer Scientist Award (2023), and has hosted a National Science and Technology Major Project.
\end{biography}

\end{sloppypar}
\end{document}